\documentclass{isprs} %
\usepackage[utf8]{inputenc}
\usepackage{subfigure}
\usepackage{setspace}
\usepackage{geometry} %
\usepackage{epstopdf}
\usepackage[labelsep=period]{caption}  %
\usepackage[british]{babel}
\usepackage{graphicx}
\usepackage{epsfig}
\usepackage{url}
\usepackage{algorithm}
\usepackage{algorithmic}
\usepackage{tikz}
\usepackage[symbol]{footmisc}

\usepackage[colorlinks=true, allcolors=blue, urlcolor=purple]{hyperref}
\usepackage{soul}%
\usepackage[colorinlistoftodos]{todonotes}
\usepackage{xargs}%
\usepackage{float}

\begin{document}
\tolerance=1000

\title{DISIR: Deep Image Segmentation with Interactive Refinement} %

\author{G. Lenczner\textsuperscript{1, 2, *}, B. Le Saux\textsuperscript{1}, N. Luminari\textsuperscript{2}, A. Chan-Hon-Tong\textsuperscript{1}, G. Le Besnerais\textsuperscript{1}}

\address{
\textsuperscript{1 }ONERA / DTIS, Universit{é} Paris-Saclay, F-91123 Palaiseau, France\\
	\textsuperscript{2 }Delair, FR-31400 Toulouse, France\\
}

\abstract{

This paper presents an interactive approach for multi-class segmentation of aerial images. Precisely, it is based on a deep neural network which exploits both RGB images and annotations.
Starting from an initial output based on the image only, our network then interactively refines this segmentation map using a concatenation of the image and user annotations. Importantly, user annotations modify the inputs of the network - not its weights - enabling a fast and smooth process. Through experiments on two public aerial datasets, we show that user annotations are extremely rewarding: each click corrects roughly 5000 pixels. We analyze the impact of different aspects of our framework such as the representation of the annotations, the volume of training data or the network architecture. 
Code is available at \href{https://github.com/delair-ai/DISIR}{this address\textsuperscript{$\dagger$}}.

}


\maketitle

\section{Introduction}\label{MANUSCRIPT}
\footnotetext[1]{Corresponding author}
\footnotetext[2]{https://github.com/delair-ai/DISIR}

Computer vision has seen tremendous progress in the last few years thanks to the emergence of powerful deep learning algorithms. This results in almost mature algorithms which are now used in industry.
However, the devil is in the details and it is often not possible to reach the precision expected by industrial end-users. To fully automate computer vision tasks, a human supervision is still often necessary to assert the quality of the results. We focus in this paper on semantic segmentation of aerial images. This task consists in image classification at the pixel level and is useful in remote sensing and Earth observation to monitor man-made architectures or natural phenomena. Using deep learning tools, it has been first addressed with fully convolutional networks in~\cite{long2015fully} and is now efficiently tackled with powerful convolutional neural networks (CNNs)
such as Deeplabv3+~\cite{chen2018encoder}.
Under appropriate conditions (e.g. when a large enough training dataset is available), one might say that semantic segmentation is nearly achieved. Indeed, these segmentation algorithms lack only a few percents of precision to reach perfect scores on public benchmarks. However, these few percents can visually make a big difference and therefore not be tolerable in practice. Besides, it often gets worse in real-life datasets due to a variety of factors (complex data, lack of well-annotated ground-truth, various usage domains, ...).  This paper proposes a fast procedure to iteratively refine the segmentation maps with a human in the loop. It consists in a neural network pre-trained with simulated human annotations and which does not require any retraining during the interactive process.

In order to concretely motivate our approach, let us consider two famous aerial image datasets in remote sensing. On the INRIA Aerial Image Labelling Dataset~\cite{maggiori2017dataset}, a building segmentation dataset, the current best networks reach an Intersection over Union (IoU) around 0.8 and a pixel accuracy around 97\% on the test set. On the ISPRS Potsdam multi-class segmentation dataset~\cite{rottensteiner2012isprs}, the state-of-the-art approaches almost reach a pixel accuracy of 92\% on the test set. While these performances are incredibly high, there might still remain some misclassified areas unacceptable for an end-user. Besides, these optimal results are obtained using top notch neural networks which have required many specific refinements~\cite{yue2019treeunet}. An off-the-shelf neural network still yields good results but, as the baselines show, a drop of performance between 5 and 10\% can be expected. Moreover, these performances decrease quickly when the networks are faced to the domain shift issues inherent to machine learning.  Therefore, the segmentation masks output by these neural networks have to be manually reviewed to meet the expectations of a potential end-user.

Let us also consider a practical application for which current approaches still yield imperfect results. Drones are increasingly used to monitor different environments like crop fields, railroads or quarries. In this context, semantic segmentation can be extremely useful for different tasks such as defects detection, volumes computation or crop monitoring. However, due to the complexity and the high variety of the acquisitions, results are usually not as good as on public datasets while a high precision is necessary for these tasks. Therefore, the operators often have to manually refine the segmentation maps which is a slow process. 

\begin{figure}[t]
   \begin{minipage}[t]{.24\linewidth}
   \centering\epsfig{figure=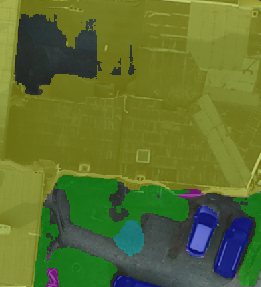,width=\linewidth}
   1 - Initial segmentation
  \end{minipage} \hfill
   \begin{minipage}[t]{.24\linewidth}
   \centering\epsfig{figure=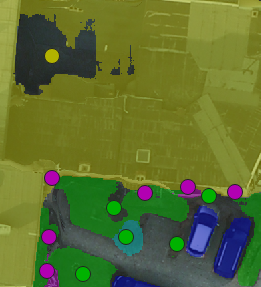,width=\linewidth}
   2 - Annotation phase
  \end{minipage} \hfill
   \begin{minipage}[t]{.24\linewidth}
   \centering\epsfig{figure=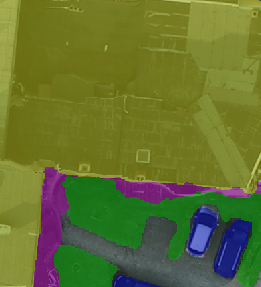,width=\linewidth}
   3 - Refined segmentation
  \end{minipage} \hfill
    \begin{minipage}[t]{.24\linewidth}
   \centering\epsfig{figure=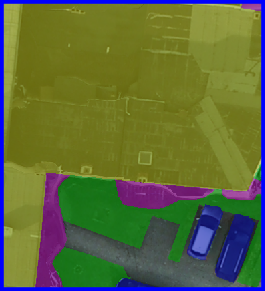,width=\linewidth}
   Ground-truth
  \end{minipage} \hfill
  \caption{Example of the proposed interactive semantic segmentation approach on the ISPRS Postdam multi-class dataset~\protect\cite{rottensteiner2012isprs}}
  \label{potsdam_segm}
 \end{figure}
To address these issues, we propose to adopt an interactive semantic segmentation approach, as sketched in Figure~\ref{potsdam_segm}. Indeed, a human in the loop can easily spot the misclassified areas and correct them thanks to a more complex yet intuitive analysis. The difficulty then is to reach optimal classification while keeping the whole process swift and engaging enough.

Our present contribution is as follows. 
\begin{enumerate}
    \item We propose an \textbf{interactive segmentation framework for aerial images using deep learning}. Once the classic training phase of a neural network is done, our algorithm does not need any retraining and can therefore quickly refine the segmentation maps.
    \item Built upon previous computer vision work mostly oriented toward interactive binary segmentation, our work focuses on interactive \textbf{multi-class segmentation} 
    \item We study the \textbf{relevance of this approach under different conditions}. Exhaustively, we investigate the impact of annotation encoding, annotation positioning, training set size and architecture choice.
\end{enumerate}

The rest of this paper is organized as follows. We first review the related work in the end of this section, second present our approach in Section~\ref{framework}, then discuss how to evaluate it in Section~\ref{eval_strat} and finally detail our experiments in Section~\ref{experiments}.
\subsection{Review of related work}
\label{review}
 \textbf{Interactive interpretation of remote sensing} data has a long history, partially due to the lack of reference data for training in that field.  Interactivity has been processed by various techniques to enhance data mining tools with relevance feedback capability~: Bayesian modeling of sample distributions was at the core of VisiMine~\cite{aksoy-interactive-KDD04}, and Support Vector Machines (SVMs) were used in~\cite{ferecatu-interactive-TGRS2007}. More recently, boosting has been the method of choice due to the possibility to train quickly in an incremental manner~\cite{dos-santos-gosselin-interactive-JSTARS2013,le_saux-14icpr-interactive}. Active learning, or in other words looking for examples which are the more able to lead to a better classification, has also been  used for this purpose~\cite{tuia2009,bruzzone-persello-2009}. With respect to these works, our approach apply deep learning for interactive remote sensing.

\textbf{Interactive segmentation} has been tackled in computer vision with a large variety of methods in the last two decades. Older ones are usually graph based methods~\cite{boykov2001interactive,rother2004grabcut,grady2006random} or based on random forests~\cite{saffari2009line,santner2009interactive}. More recently, best segmentation performances were obtained with CNN-based architectures. So, they are favored to provide the initial segmentation. Several works have then tried to make them interactive to get finer results. They standardly use points resulting from user clicks as annotations. We now review thoroughly these methods.%

Deep interactive object selection (DIOS)~\cite{xu2016deep} is the first proposal of an interactive segmentation framework based on neural networks. It aims for binary classification. In a nutshell, the network takes as input two additional channels concatenated with the RGB image. The first one contains annotation points from the foreground while the other one contains background points. These annotation points are encoded into euclidean distance maps. The annotations are automatically sampled during training using the ground-truth maps.  
We extend this approach to multi-class segmentation of aerial images. %
Multiple existing works are inspired by DIOS. 
\cite{liew2017regional} adopt a multi-scale strategy which refines the global prediction by combining it with local patch-based classification. %
\cite{hu2019fully} also follow a multi-scale strategy by designing a two-stream fusion network to process the annotations differently than the image.%
 A particular challenge is to get enough useful annotations. For this purpose,~\cite{mahadevan2018iteratively} use a hard-sample mining strategy at training by selecting annotations among erroneous predictions.%
Alternatively,~\cite{jang2019interactive} iteratively optimize the annotation maps given as inputs by back-propagating the errors between predictions and annotations. %
Finally, DEXTR~\cite{maninis2018deep} and~\cite{wang2019object} both ask the user to click points on the borders and corners of the objects. %
Recently,~\cite{benenson2019large} assess these various strategies in the first large scale study of interactive instance segmentation with human annotators. Their experiments hint that center annotation clicks are the most robust and that distance transform to encode the annotation points can be replaced by binary disks.

Polygon-RNN++~\cite{acuna2018efficient} is  an interesting alternative to DIOS-like approaches. Using a CNN-RNN architecture, they predict a polygon which can be refined by moving its vertices. Using Graph
Convolutional Networks (GCN), Curve-GCN~\cite{ling2019fast} extend this work by predicting a spline which better outlines curved objects. Note that these aforementioned approaches aim to binary classification.%

\textbf{Multi-class interactive segmentation} has also been approached in various ways. Several older methods~\cite{nieuwenhuis2014co,nieuwenhuis2012spatially} address this problem using a bayesian maximum a posteriori (MAP) approach while~\cite{santner2010interactive} rely on a random forest classifier. Recently,~\cite{andriluka2018fluid} use a combination of two slightly modified Mask-RCNN~\cite{He_2017_ICCV} to compute multiple fixed segmentation propositions and then let the user choose which of these propositions should form the final segmentation. Finally, \cite{agustsson2019interactive} are the first to propose a deep learning approach which lets the user correct the shape of a proposed multi-class segmentation. Their algorithm takes as input a concatenation of the image and the extreme points of each instance in the scene and then corrects the segmentation proposal using scribbles. In contrast, we adopt in our work a class-dependant DIOS-based approach to refine an initial segmentation map. 

\textbf{Automatic evaluation of an interactive system} requires some way of mimicking the human annotation. \cite{xu2016deep} try to mimick a user who would correct the largest mislabelled regions by iteratively placing the annotations far from the boundaries of the mislabelled prediction. In details, they use as annotation the point which maximises the distance to the boundary of the false prediction in order to simulate an image analyst. Then, they do a new prediction using this generated annotation and repeat this process for 20 iterations. Our automatic clicking strategy is inspired from this idea and adapted to our multi-label segmentation problem. \cite{benenson2019large} add some noise in the simulated clicks to better match the human behavior. They also manually experiment their approach with a large pool of human annotators. In~\ref{eval_strat}, we investigate different strategies to simulate human analysts and evaluate automatically an interactive system.

\section{Proposed algorithm}

\begin{figure*}[h!]
\begin{minipage}{.49\linewidth}
\centering\epsfig{figure=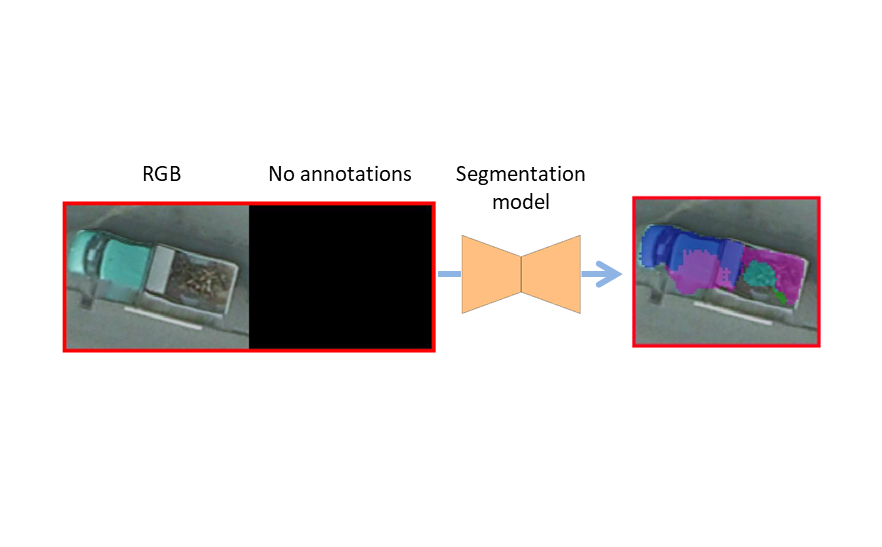,width=\linewidth}
Initialization phase
\end{minipage}\hfill
\begin{minipage}{.49\linewidth}
\centering\epsfig{figure=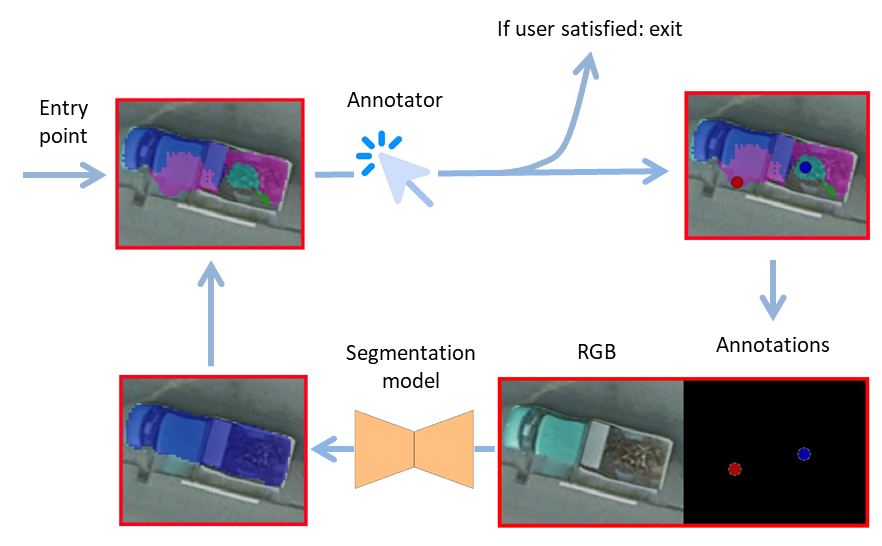,width=\linewidth}
Interactive loop
\end{minipage}\hfill
\caption{High level overview of the proposed approach: information provided by the user modifies the input of the network - not the network itself - allowing an effective interaction}
\label{high_overview}
 \end{figure*}

\label{framework}

We now describe in details the proposed approach for interactive multi-class segmentation of aerial images. In particular, our goal is to train a neural network with two purposes:
\begin{enumerate}
    \item producing an initial high quality segmentation map of the scene without any external help;
    \item using annotations provided by an operator to quickly enhance its initial prediction.
\end{enumerate}

To achieve this, we propose a neural network which keeps its original structure but takes as input a concatenation of the classic inputs (e.g. RGB) and of the annotations ($N$ channels, one per class). These annotations are clicked points. Note that only the inputs of the network are modified and not its weights: this makes the swiftness of the approach. Figure~\ref{high_overview} presents a high-level overview of our approach.

We first define our training strategy and then present our study on the annotations themselves. %
\subsection{Training strategy}

In the following, we assume that we have a segmentation reference composed of $N$ classes. Ground-truth maps are the core of our training strategy. On one hand, they are classically used to compute and back-propagate the loss. On the other hand, they are also \textit{randomly sparsified to sample annotations}. In other words, only a few pixels from the ground-truth are kept to be used as annotations. According to their class, these annotations are encoded in the $N$ annotation channels given as input to the algorithm. %
 To train under various annotation layouts, the number of sampled annotations is random in each training example. Since the network has to be able to create an accurate segmentation map without them, the possibility of a lack of annotations is also sampled. Concretely, this situation means that the annotation channels are filled with zeros.

If the annotations are sampled independently of their class, the following problem may occur. During the evaluation phase, annotations on sub-represented classes can be ignored by the network because it has barely seen any annotation points of these classes during training. Therefore, it has not learned how to use them to enhance its predictions. To overcome this issue, we use a \textit{frequency balancing} strategy to sample the annotations based on the classes distributions. It allows the network to equally see annotations from each class during training and, therefore, to be efficiently guided once the training is done.  

\subsection{Annotation representation}%
\label{annots_study}%

We investigate two aspects of the annotation representation: how to \textit{position} clicks in order to sample the most useful information, and how to \textit{encode} clicks to get the best benefit.  

\paragraph{Click positioning.}
Fixing a wrong segmentation implies to provide the system with additional information about the right division. New samples provided by clicks may represent either the inside of an instance or its border.

The first case seems to be the most intuitive. Clicked pixels are inside instances and the annotation points represent the class associated to these instances. Contrary to~\cite{xu2016deep}, we do not sample them at a minimal distance from the boundaries since we assume that an annotator might click near an edge to fine-tune the prediction. For the second case where the annotations represent the borders of the instances, the channel associated to a click corresponds to a class randomly chosen among the ones adjacent to the clicked border. %

Aiming to ease the burden of the end users, we also explored softer constraints on the annotations. Indeed, instead of using $N$ annotation channels, we summarized them into a single annotation channel. For the border strategy, this single channel only indicates the presence of a border. For the inside point strategy, it only indicates where the network has initially made a mistake. To implement this latter strategy, we had to slightly modify the training process. The network performs a first inference to create a segmentation map used to find mislabelled regions. Annotations are then sampled in these areas and a second inference is performed. Only this second inference is used to back-propagate the gradients.
However, as shown in Section~\ref{infl_annot_strat}, none of these simplified annotations seems promising to efficiently guide the segmentation task.

\begin{figure}
  \centering\epsfig{figure=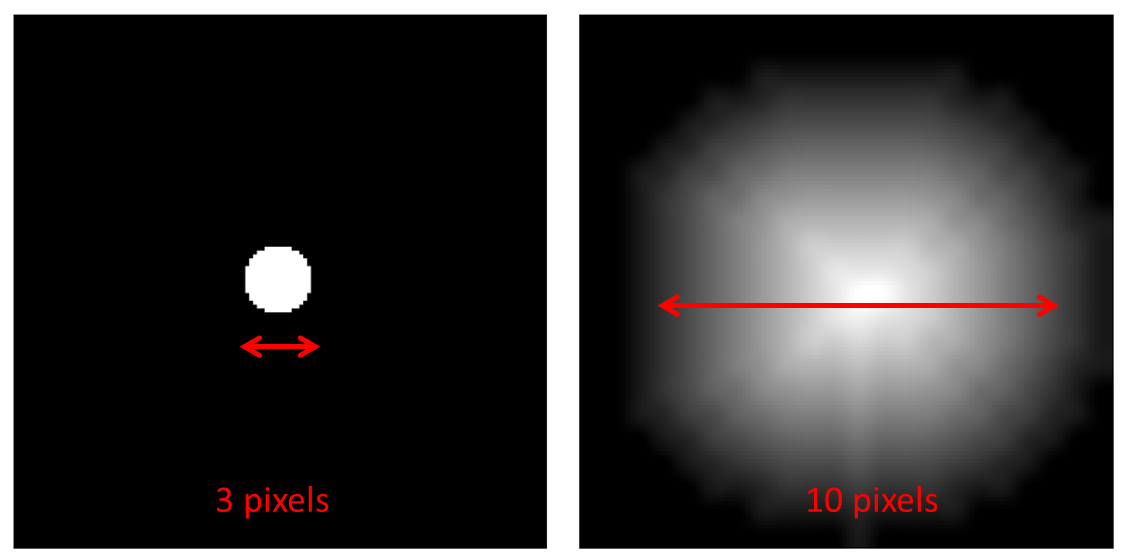,width=\linewidth}
 \caption{Binary (left) and distance transform (right) click.}
 \label{clicks_encoding}
\end{figure}

\paragraph{Click encoding.}
User clicks can be encoded in various ways, and such may provide the system with more or less spatial information, as shown in Figure~\ref{clicks_encoding}. In particular, we consider:

\begin{itemize}
    \item Small binary area around the annotation points
    \item Euclidean distance transform maps around these points
\end{itemize}

As shown in Section~\ref{experiments}, the inside point strategy with distance transform encoding seems to be our most successful combination.

\section{Discussion on the evaluation strategy}
\label{eval_strat}
To evaluate interactive approaches, two procedures are possible. Either the ground-truth is available and can be used to sample annotations, or it is not and a human operator has to make the annotations. Therefore, we have evaluated our approach both manually and automatically. We use the IoU averaged over all classes as our evaluation metric.%
\paragraph{Automatic evaluation}
For each image, the neural network makes a first inference without annotations. A click is then automatically generated and the network makes a new inference. This process is repeated iteratively for a fixed number of clicks.

We have compared two click sampling strategies, both using a comparison between the prediction and the ground-truth:
\begin{enumerate}
    \item A click is automatically sampled in one of the biggest mislabelled areas. Some randomness is added in the choice of the area and in the localization of the click inside the area to better simulate a human behavior. This process is class-independent.
    \item  The process is similar to 1. but the generated click has the supplementary constraint to be on a pixel belonging to a specified class. This allows to also correct pixels belonging to sub-represented classes. This process is class-dependent.
\end{enumerate}

As we will see in~\ref{auto_eval}, the class-dependent process is better than the first one to evaluate the influence of the clicks on sub-represented classes but has two drawbacks: overall smaller erroneous areas are corrected which leads to smaller corrections and there is less room for randomness if the chosen class to annotate is predetermined.

\paragraph{Manual evaluation}
This process is similar to the automatic one but the clicks are now made by a human operator. This operator also aims to correct the biggest erroneous areas but the localization of the clicks is now inherently subjective. %

To do this manual evaluation, we have built a QGIS~\cite{QGIS_software} plugin. User interaction is then handled by the QGIS interface while
the heavy computations, e.g. the semantic segmentation, are performed in a separate server that can be local or remote. Once the server is launched, the data transfer is transparent to the user.

\section{Experiments}
\label{experiments}
In this section, we aim to show that our method works and how to best evaluate it among the two evaluation strategies aforementioned in Section~\ref{eval_strat}. Besides, we study the influence of the different parameters described in Section~\ref{annots_study}. Furthermore, we conduct two experiments to better apprehend the possibilities and the limits of our approach:
\begin{itemize}
    \item We have first compared different backbone architectures to evaluate if it has a significant impact on the performances. More importantly, since these different architectures produce different initial segmentation maps, this comparison also allows us to study if the initial quality of the segmentation maps influences the benefits brought by the annotations.%
    \item The second one is motivated by the fact that it often happens in practice to only have access to a very limited amount of annotated data. Similar to~\cite{castillo2019data} where the authors study the influence of the training set size on the network performance, we study the influence of this parameter on the neural network refinement abilities. To this end, we have trained the networks on subsets of the initial training sets. %
\end{itemize}

This section is thus organized as follows. We first present our experimental setup. Second, we show that our method works and how to best evaluate it. Then, we analyse the outcomes of our different experiments with automatic evaluation. Finally, we draw conclusions from the manual evaluations. 
\subsection{Experimental setup}

\paragraph{Datasets.}We have tested our approach on the two standard remote sensing datasets mentionned in the Introduction.
We split the initial training sets into a training and a validation sets with a 80\%-20\% ratio and use the validation sets for our experiments. The INRIA Aerial Image Labelling dataset~\cite{maggiori2017dataset} is composed of two classes (\textit{buildings} and \textit{not buildings}) and covers more than 800 $\textrm{km}^2$ with a spatial resolution of 0.3m. The size of each image is $5000\times5000$ pixels. The training set is composed of 144 images and the validation set of 36 images.
 The ISPRS Potsdam dataset~\cite{rottensteiner2012isprs} is composed of 6 classes (\textit{impervious surface, buildings, low vegetation, tree, car} and \textit{clutter}). The class \textit{car} is sub-represented compared to the other classes. This dataset covers around 3~$\textrm{km}^2$ with a spatial resolution of 0.05m. The size of each image is $6000\times6000$ pixels. The training set is composed of 19 images and the validation set of 5 images.

\paragraph{Neural Network.}Except in the backbone comparison, we use a LinkNet~\cite{chaurasia2017linknet} architecture. It is a classic encoder/decoder architecture relying on a ResNet encoder~\cite{he2016deep}. The networks are trained using stochastic gradient descent (SGD) and cross-entropy loss for 50 epochs with a batch of size 8, seeing during each epoch 10000 samples randomly chosen and cropped (size $512\times512$). The initial learning rate is fixed at 0.05 and is divided by 10 after 15, 30 and 45 epochs. Only basic data augmentation is performed: horizontal and vertical flips. The implementation is done using Pytorch.%

\paragraph{Annotations.}During our different evaluations, we sample 120 clicks for each image and measure the IoU gain for each class. 
Except when specified otherwise, we encode clicks using distance transform and assume that they represent the inside of the corrected instances. The annotations for the Potsdam dataset are sampled during training based on their class distribution, except in the frequency balancing influence study. %

\subsection{Approach assessment and automatic evaluation strategy}
\label{auto_eval}
\begin{figure}[h!]
  \begin{minipage}{1\linewidth}
   \centering\epsfig{figure=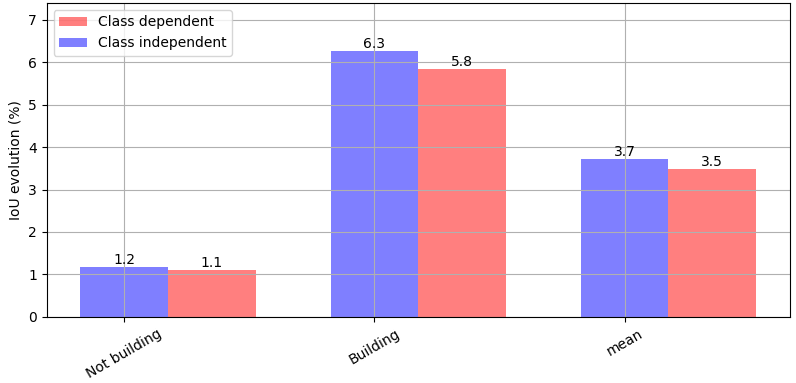,width=\linewidth}
  \end{minipage}
  \begin{minipage}{1\linewidth}

   \centering\epsfig{figure=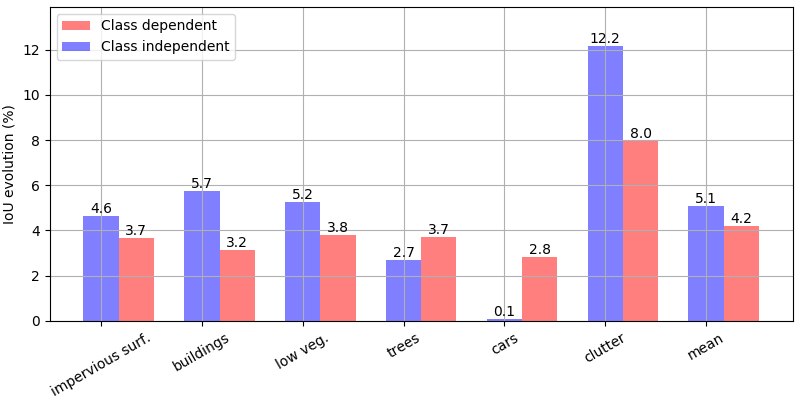,width=\linewidth}
  \end{minipage}
  \caption{Comparison of the two different automatic evaluation processes on the INRIA (top) and Potsdam (bottom) datasets.}
  \label{fig:auto_eval}
 \end{figure}

Here we compare the two proposed automatic 
evaluation strategies. Let us recall that, for the first strategy, the clicks are sampled iteratively in the biggest erroneous areas independently of the class while they are sampled equally in each class in the second one. 
As we can see on Figure~\ref{fig:auto_eval}, the class independent evaluation strategy allows to reach a higher overall IoU but to the detriment of the sub-represented \textit{car} class in the Potsdam dataset. This is due to the fact that the biggest erroneously predicted areas inherently belong to larger instances than cars such as buildings. Therefore, even though the overall metric gain is not as good as with the class-independent evaluation process, we choose the class-dependent one for our further evaluations on Potsdam. However, the INRIA dataset does not contain a low-represented class so we choose the class-independent evaluation process to evaluate our experiments on this dataset.

Whatever the evaluation strategy, the results displayed in Figure~\ref{fig:auto_eval} validate the efficiency of our approach on both datasets. Indeed, all segmentation performances are improved for all classes: on average the mean IoU is increased by $3.7\%$ on the INRIA Building dataset, and by $4.2\%$ on the muti-class ISPRS Potsdam dataset. Besides, as we can see on Table~\ref{tab:tab1}, each click allows to correct around $5000$ pixels in average.
\begin{table}[]
    \centering
    \begin{tabular}{c|c}
        Dataset & Corrected pixels\\ \hline
        INRIA & 3143\\
        Potsdam & 7219
    \end{tabular}
    \caption{Average corrected pixels per click}
    \label{tab:tab1}
\end{table}

\subsection{Influence of the frequency balancing}

\begin{figure}[h!]
\begin{center}
		\includegraphics[width=1.0\columnwidth]{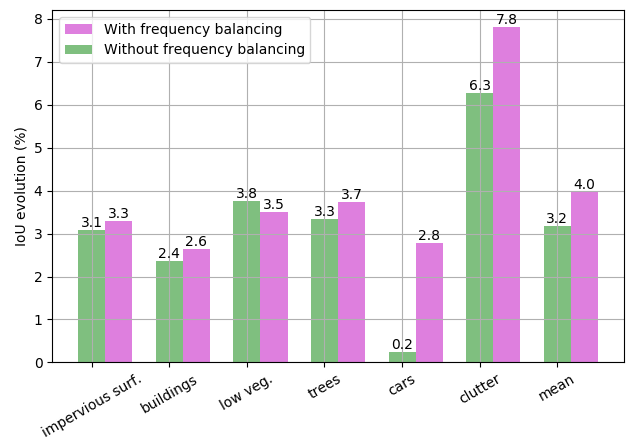}
	\caption{Study of the impact of the frequency balancing during training on the Potsdam validation set.} %
\label{fig:freq}
\end{center}
\end{figure}

Figure~\ref{fig:freq} compares the improvements between training with frequency balancing to sample the annotations and without. Here, we use binary-encoded clicks in order to control the number of samples per class. As we can see, without frequency balancing, the network does not learn to use \textit{car} annotations to refine its predictions. Indeed, the low-representation of this class implies that only few annotations from it were seen during training. A frequency balancing strategy efficiently tackles this issue. Overall, mean IoU is increased by almost $1\%$ and 5 classes out of 6 are improved. However, frequency balancing is not necessary for the INRIA dataset which does not contain any low-represented class.

\subsection{Influence of the annotation strategy}
\label{infl_annot_strat}
\begin{figure}[h!]
  \begin{minipage}{1\linewidth}
   \centering\epsfig{figure=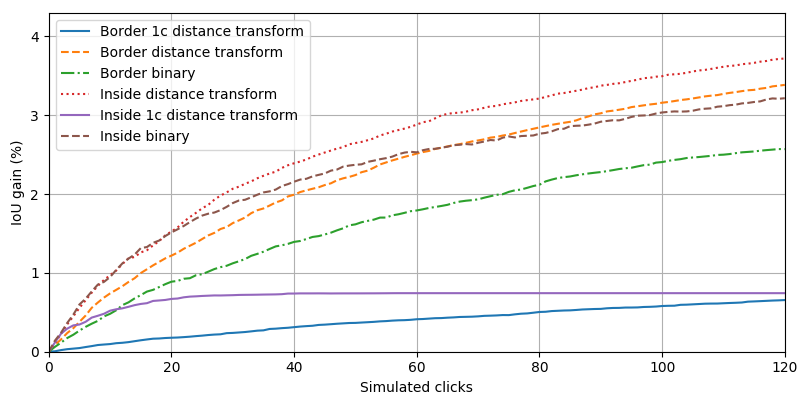,width=\linewidth}
  \end{minipage}
  \begin{minipage}{1\linewidth}

   \centering\epsfig{figure=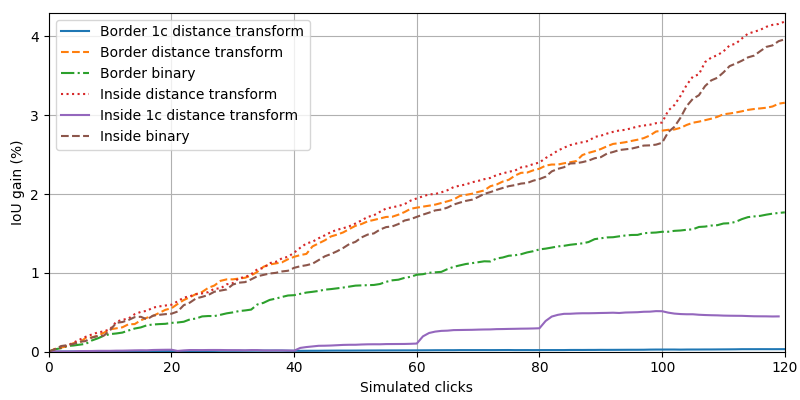,width=\linewidth}
  \end{minipage}
  \caption{Comparison of the different annotation strategies on the INRIA (top) and Potsdam (bottom) datasets. \textit{1c} means single annotation channel.}
  \label{fig:annot_strat}
 \end{figure}

We compare here the different annotation strategies: inside clicks or border clicks, binary encoding or distance transform, and single or multiple channels. As we can see on Figure~\ref{fig:annot_strat}, the distance transform clearly increases the benefits of the annotations compared to the binary encoding.
While~\cite{benenson2019large} conclude that the binary encoding leads to better performances, our opposite conclusion might be inherent to the large size and scale of aerial images
which dilute the annotations localized over very small areas.

Both the contours and the inside points are efficiently used by the network to enhance its predictions but it is still noticeably better with the inside points. We can also notice that the last 20 added points considerably boost the performances of the inside point strategies for the Potsdam dataset: this is due to the fact that these points belong to the class \textit{clutter}, an under-represented class. Therefore, they have the strongest impact in term of IoU.
Finally, the two degraded strategies which rely on single annotation channel bring little or no improvement even though it is slightly better for the INRIA dataset since it contains only two classes.  

\subsection{Influence of the network backbone}
\begin{figure}[h!]
\begin{center}
		\includegraphics[width=1\columnwidth]{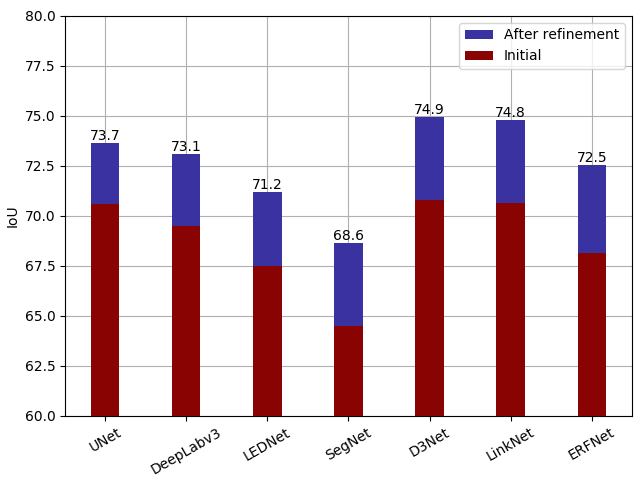}
	\caption{Study of the impact of the architecture choice on the Potsdam validation set sorted per IoU gain.}
\label{fig:diff_archis}
\end{center}
\end{figure}

We compare LinkNet to SegNet \cite{badrinarayanan2017segnet}, UNet~\cite{ronneberger2015u} and DeepLabv3~\cite{chen2017rethinking} which are standard segmentation networks with increasing complexity and also to the following lighter architectures: LEDNet~\cite{wang2019lednet}, ERFNet~\cite{romera2017erfnet} and D3Net~\cite{carvalho2018regression}.
Figure~\ref{fig:diff_archis} shows the results obtained with the different architectures under the same training and evaluating conditions. As expected since this framework is agnostic to the network architecture, the gains are in the same order of magnitude. Indeed, the initial IoU mean is $68.8\%$ with a standard deviation of 2.13 while the IoU gain mean is $3.9\%$ with a standard deviation of 0.4. Figure~\ref{fig:diff_archis} also shows that the accuracy gain of the interactive correction seems to be uncorrelated to the accuracy of the initial segmentation map. For instance, the worse initial architecture here -- SegNet -- is the average one in regards to the IoU gain.

\subsection{Influence of the volume of training data}
\begin{figure}[h!]
\begin{center}
		\includegraphics[width=1\columnwidth]{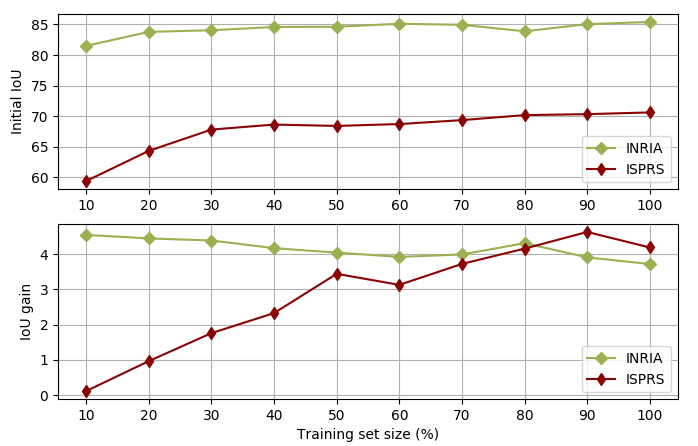}
	\caption{Influence of the training set size on the initial IoU (top) and on the IoU gain (bottom)}
\label{fig:subset}
\end{center}
\end{figure}
Figure~\ref{fig:subset} shows the influence of the training set size on our approach. The different behavior on the two datasets can be explained by their initial size difference.

On the INRIA dataset, since the initial training size is high, even 10\% of the training data seems to be enough to provide a network with a decent initial accuracy and a good ability to use the annotations. Besides, if more data implies a better initial accuracy, it does not improve the performances of the interactive correction. This shows that the network has not learned to make a better use of the annotations with supplementary data. %

On the Potsdam dataset, even though the initial training size is lower than in the INRIA dataset, the network is still initially quite accurate with little training data. Indeed, according to the results of~\cite{castillo2019data}, since there are pictures from only one city, few training images are enough to learn the general semantic of the dataset even if the full training set provides better performances. However, the accuracy gain is really low with little training data. For example, the IoU gain is less than 1\% with 20\% of the initial volume of data while it is slightly over 4\% with the full training set. We believe that this lack of performance in low-data regime is due to over-fitting. Indeed, since there are only a few training images in this scenario, there are also less possible annotations and they might not fully reflect the reality of the test set.
Besides, if the network over-fits on these few images, it might also consider the annotations as unnecessary for the segmentation. 

Therefore, as shown by the study on the Potsdam dataset, a certain amount of data seems necessary to optimally use the annotations. However, as shown by the study on the INRIA dataset, the network ability to use the annotations reaches a plateau once there is enough available training data.

\subsection{Manual analysis}

In this experiment, the images from the Potsdam validation set have been manually refined by a human annotator. If the number of clicks exceeds 120, we threshold it at 120 in order to make a fair comparison with the the automatic process.

\begin{figure}[h]
   \begin{minipage}[t]{.24\linewidth}
   \centering\epsfig{figure=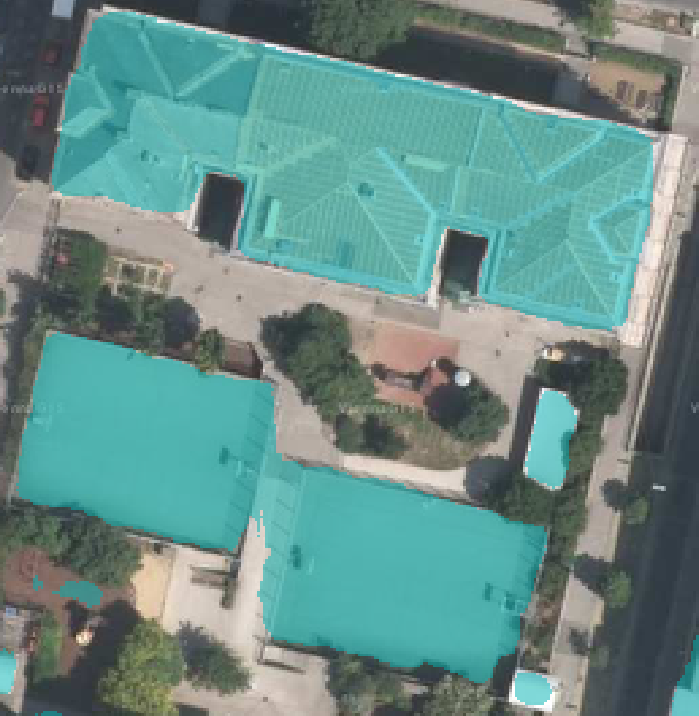,width=\linewidth}
   1 - Initial segmentation
  \end{minipage} \hfill
   \begin{minipage}[t]{.24\linewidth}
   \centering\epsfig{figure=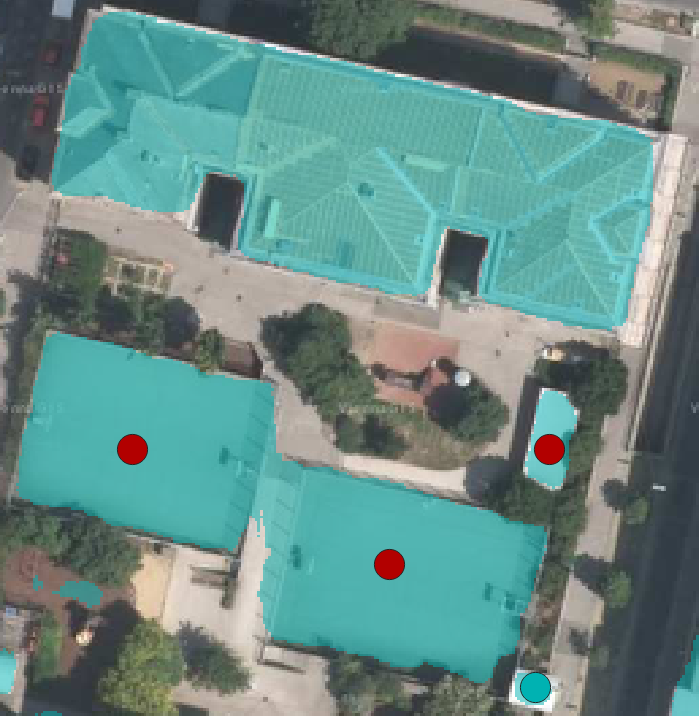,width=\linewidth}
   2 - Annotation phase
  \end{minipage} \hfill
   \begin{minipage}[t]{.24\linewidth}
   \centering\epsfig{figure=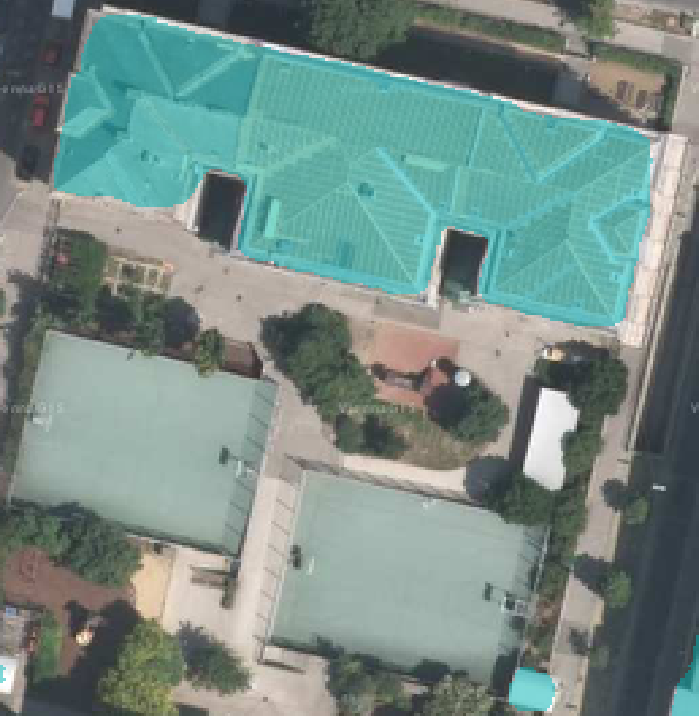,width=\linewidth}
   3 - Refined segmentation
  \end{minipage} \hfill
    \begin{minipage}[t]{.24\linewidth}
   \centering\epsfig{figure=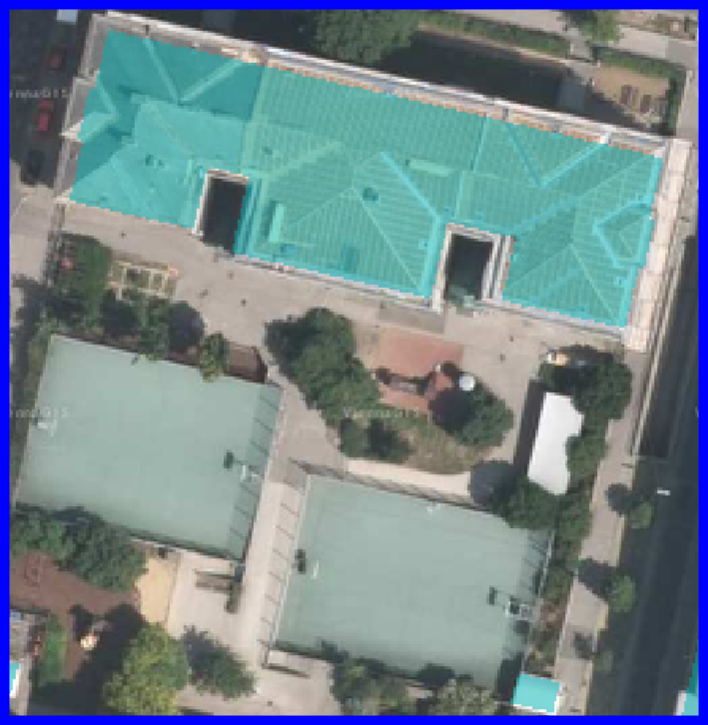,width=\linewidth}
   Ground-truth
  \end{minipage} \hfill
  \caption{Annotations lead to an easy false positive buildings removal on the segmentation map}
  \label{buildings_segm}
 \end{figure}

\paragraph{Local insights.} On one hand, as shown on Figure~\ref{buildings_segm}, the refinements can be very intuitive and effective on areas semantically similar to the ones seen during training. On the other hand, if the semantic is new compared to what is in the training set, the neural networks have trouble to use the annotations efficiently.%

  \begin{figure}[h!]
   \begin{minipage}[t]{.47\linewidth}
   \centering\epsfig{figure=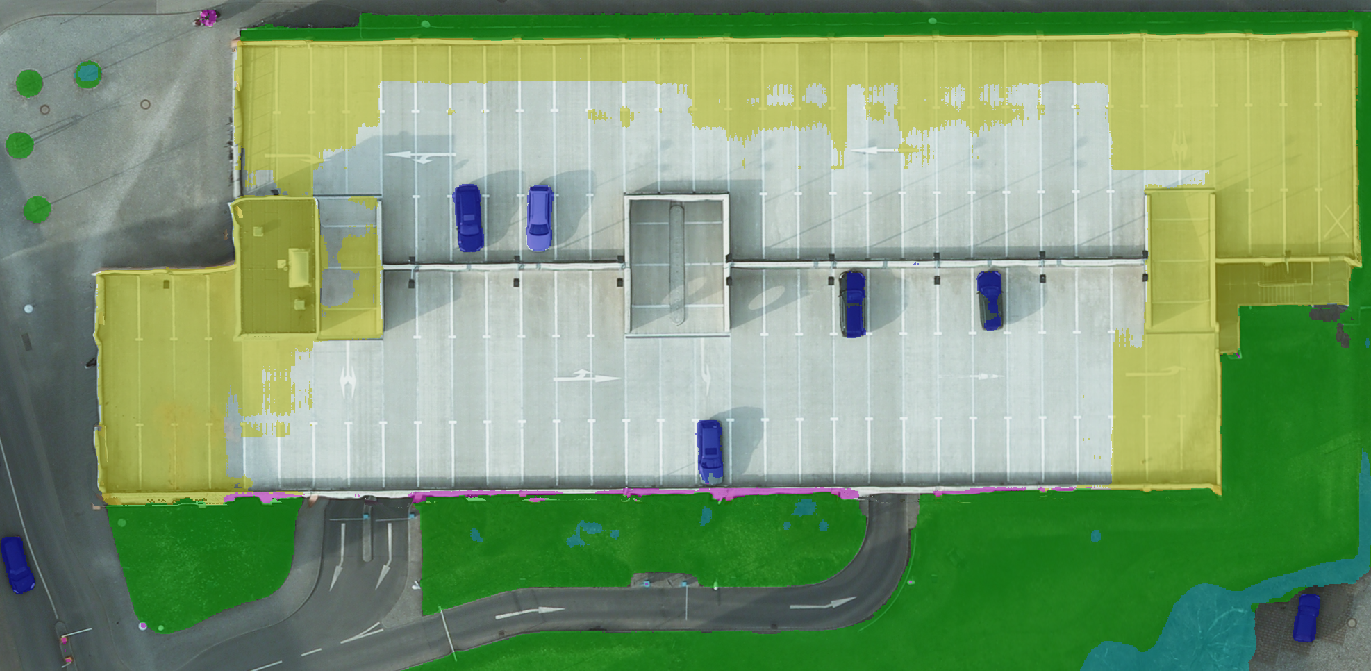,width=\linewidth}
   1 - Initial segmentation
  \end{minipage} \hfill
   \begin{minipage}[t]{.47\linewidth}
   \centering\epsfig{figure=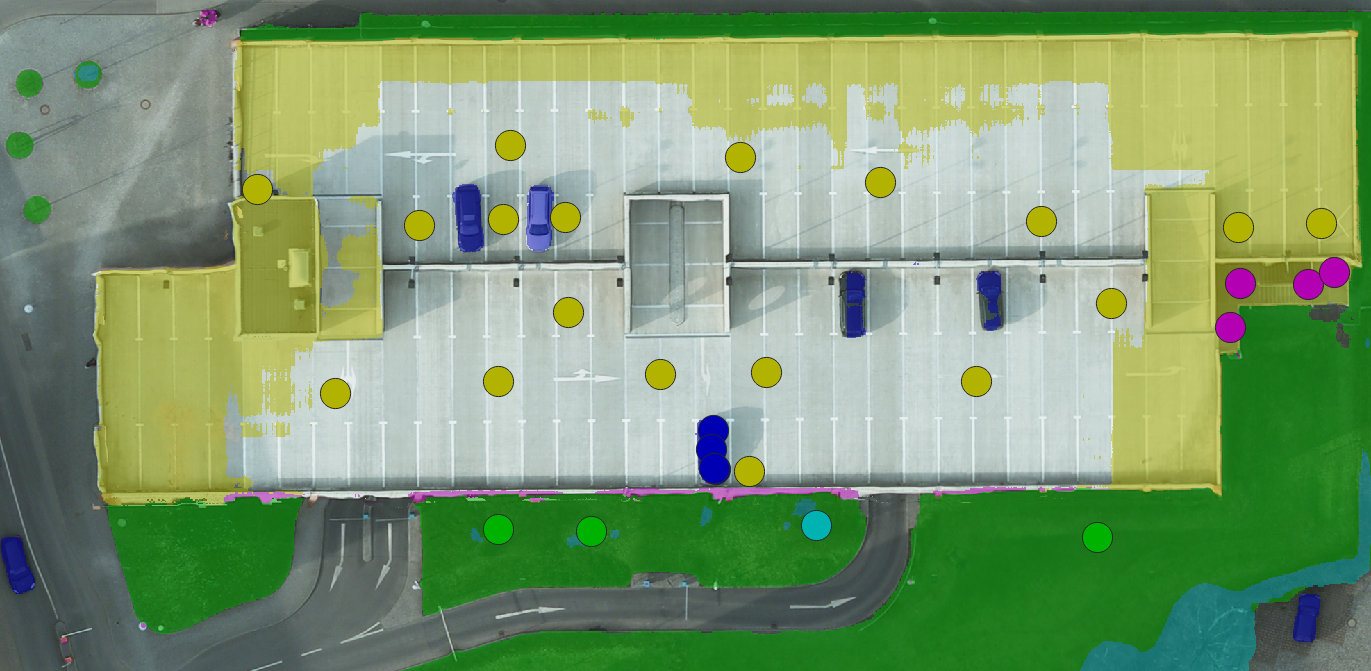,width=\linewidth}
   2 - Annotation phase
  \end{minipage} \hfill
   \begin{minipage}[t]{.47\linewidth}
   \centering\epsfig{figure=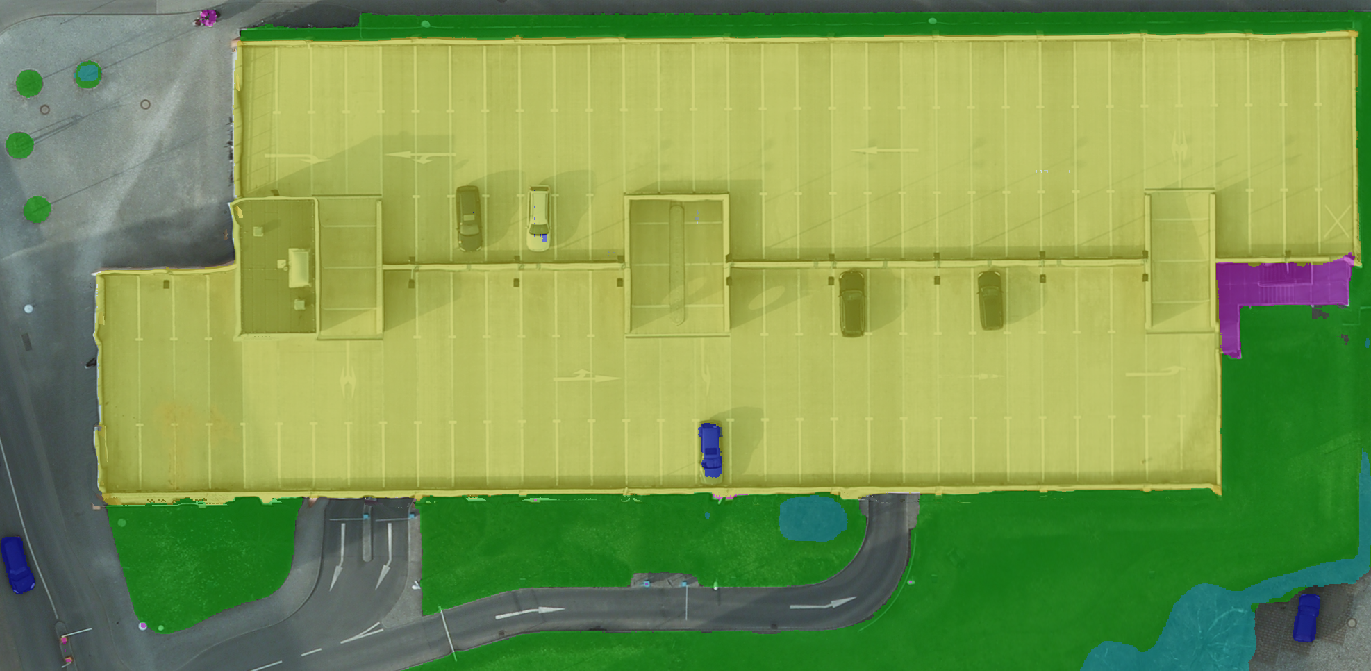,width=\linewidth}
   3 - Refined segmentation
  \end{minipage} \hfill
    \begin{minipage}[t]{.47\linewidth}
   \centering\epsfig{figure=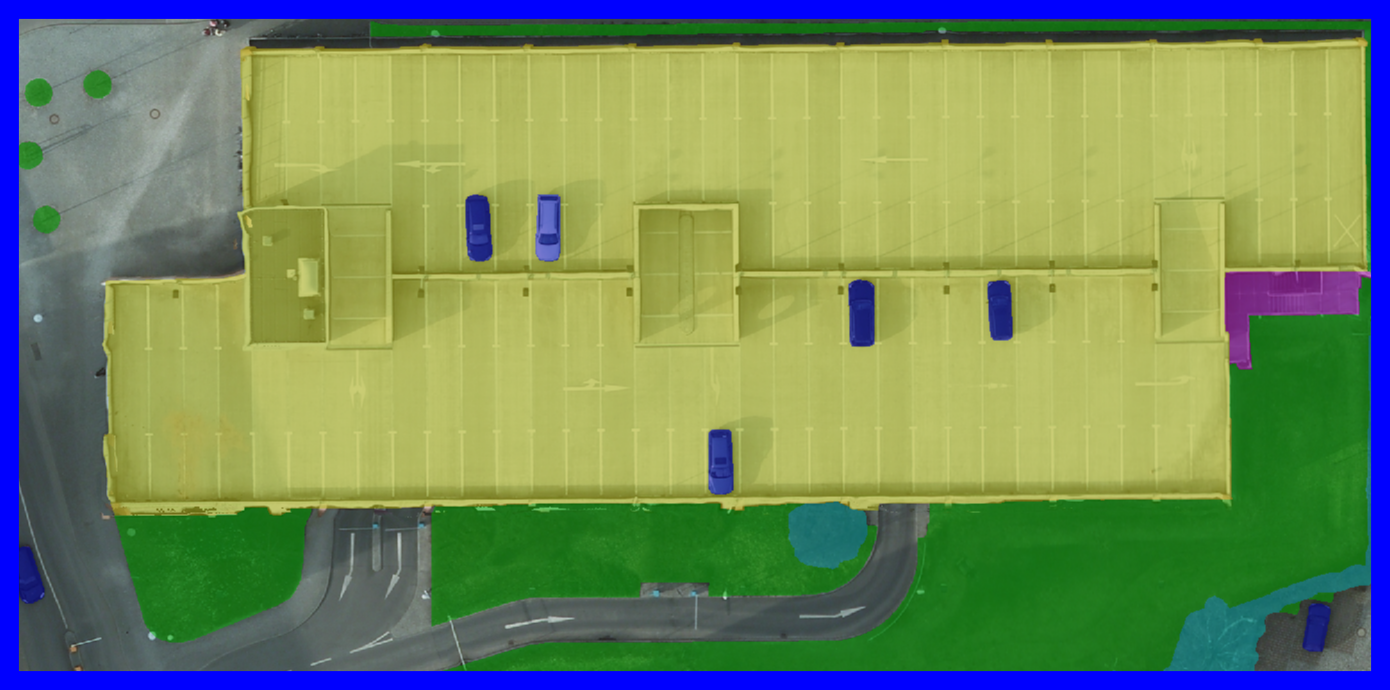,width=\linewidth}
   Ground-truth
  \end{minipage} \hfill
  \caption{Difficult segmentation of an outside parking since the network has not learned the semantic \textit{cars on building}. Only the car at the bottom is annotated and recognized.} %
  \label{cars_on_build}
 \end{figure}
For example, in the Potsdam dataset, there is only one outside car park considered as building which means only one place with the semantic "\textit{car}" surrounded by "\textit{building}" in the dataset. We kept the associated image in the validation set to study the impact of the annotations in this scenario. Figure~\ref{cars_on_build} shows the outcome of our approach on this car park. Since it also looks like a road, it is initially difficult for the network to segment it correctly. Nonetheless, it succeeds to recognize the cars parked there. Then, with building annotations, the network successfully recognizes a building. However, it also considers that the vehicles parked there are now part of the building since it has never seen the class "\textit{car}" surrounded by the class "\textit{building}" during training. As we can see on Figure~\ref{cars_on_build}, with supplementary \textit{car} annotations, the network can still recognize the correct semantic of the scene. However, the process in this case is not very smooth and intuitive since the cars which were primarily well recognized need to be annotated nonetheless. 
This example shows that our framework does not perform optimally when it is faced to areas with a different semantic compared to the ones present in the training set.%

\begin{figure}[h!]
\begin{center}
		\includegraphics[width=1.0\columnwidth]{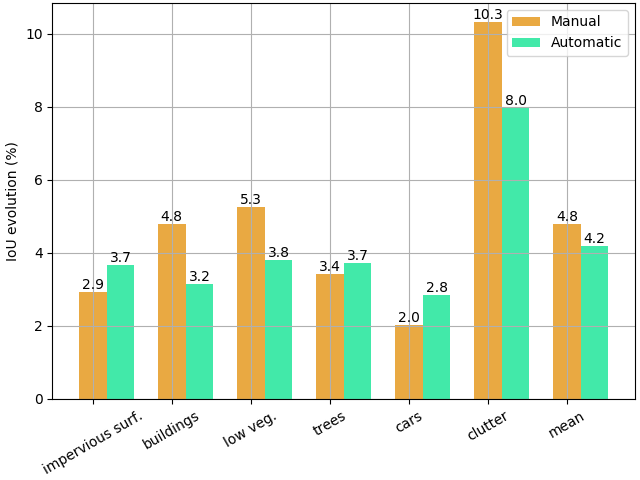}
	\caption{Comparison of the IoU evolution between an automatic and a manual evaluation on the Potsdam dataset.} %
\label{fig:automanu}
\end{center}
\end{figure}

\begin{figure}[h!]
    \begin{minipage}{.32\linewidth}
    \centering\epsfig{figure=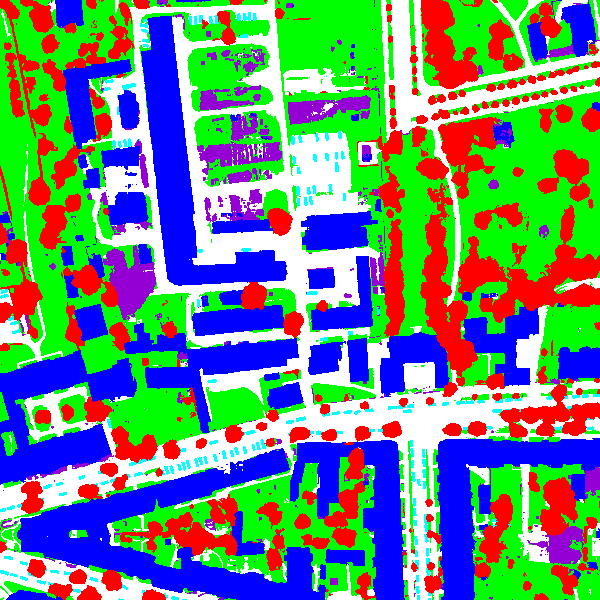,width=\linewidth}
    Before HI (87.5\%)
  \end{minipage}\hfill
  \begin{minipage}{.32\linewidth}
   \centering\epsfig{figure=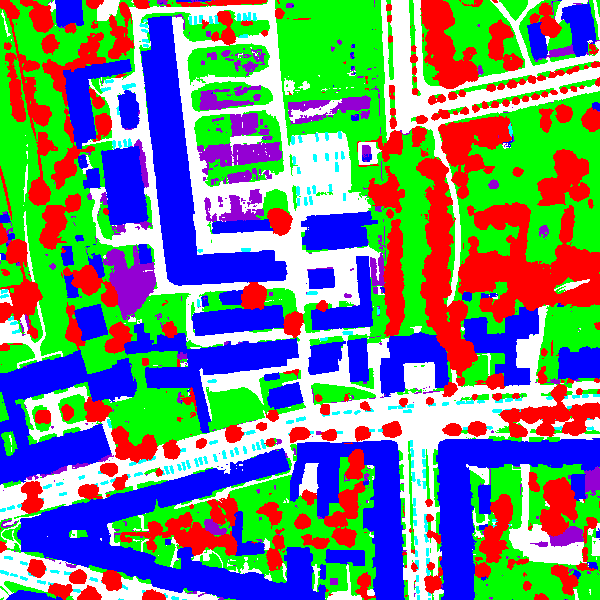,width=\linewidth}
   After HI (90.4\%)
  \end{minipage}\hfill
    \begin{minipage}{.32\linewidth}
  \centering\epsfig{figure=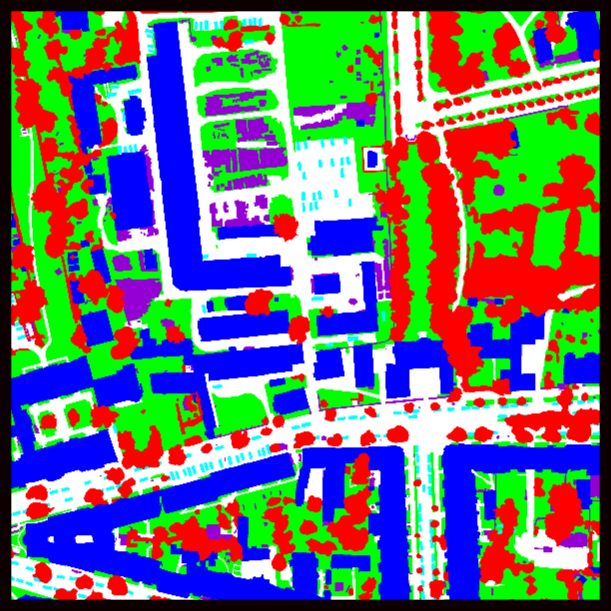,width=\linewidth}
  Ground-truth
  \end{minipage}\hfill
  \caption{Full predictions and their accuracy before/after human interaction (HI) on an image from Potsdam}
  \label{fig:full_preds}
 \end{figure}

\paragraph{General insights.}
\begin{figure}[h!]
  \begin{minipage}{.48\linewidth}
   \centering\epsfig{figure=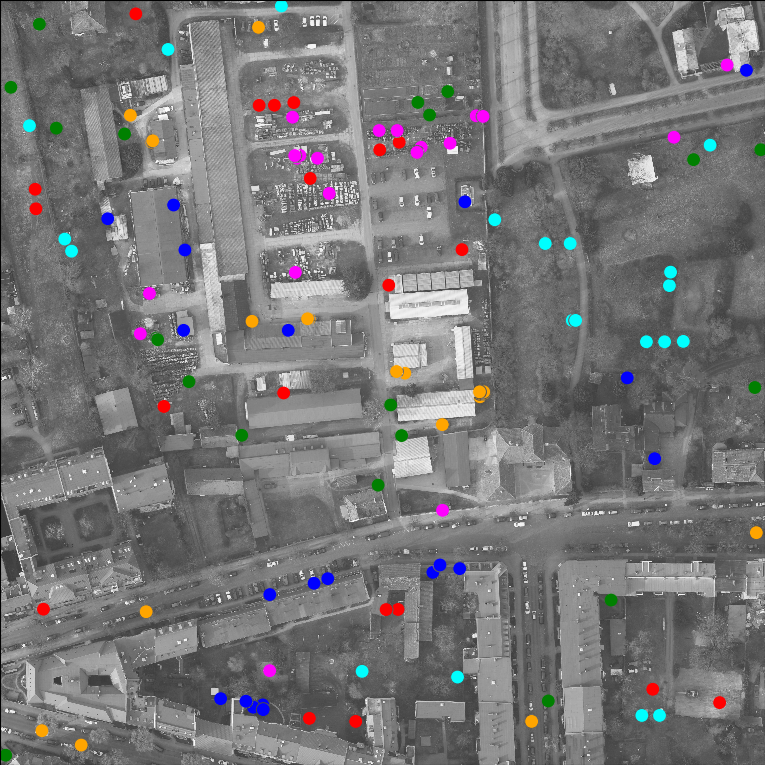,width=\linewidth}
  \end{minipage}\hfill
  \begin{minipage}{.48\linewidth}

   \centering\epsfig{figure=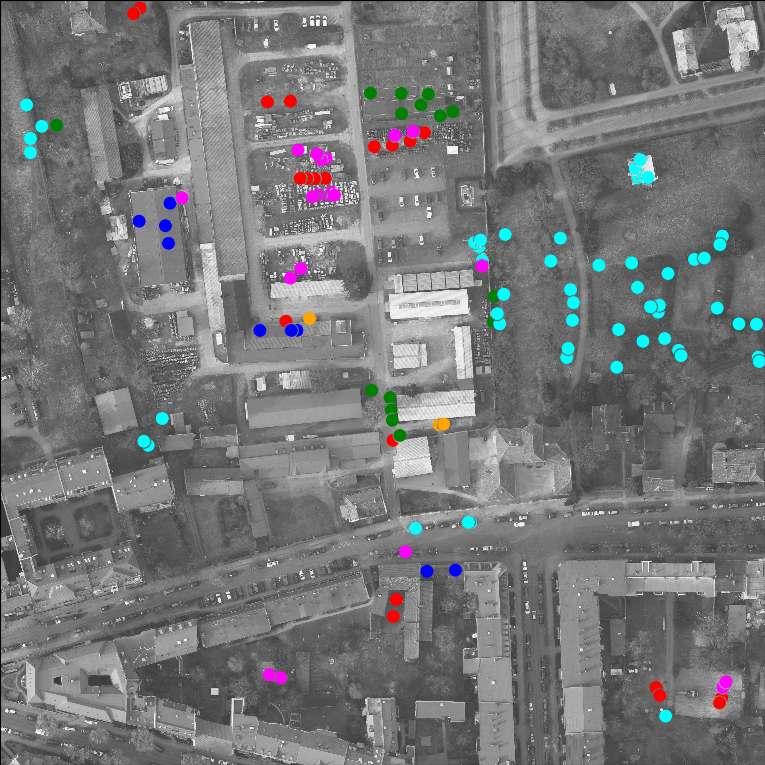,width=\linewidth}
  \end{minipage}\hfill
  \caption{Click distribution on an image from Potsdam in automatic (left) and manual (right) evaluations. The colors represent the different classes.}
  \label{fig:clicks_comp}
 \end{figure}
 Regarding the click distribution, as shown in Figure~\ref{fig:clicks_comp}, a human operator tends to focus clicking on specific areas while the automatic evaluation rather spreads the annotations all across the image. However, as shown in Figure~\ref{fig:automanu}, these grouped clicks seem to efficiently increase the metric. Indeed, with the manual evaluation, 4 classes out of 6 are more improved and the mean IoU gain is overall better. This shows the efficiency of our approach with a real user in the loop. Finally, Figure~\ref{fig:full_preds} shows qualitative results before and after human interaction. 

\section{Conclusion}
We have proposed in this article an interactive multi-class segmentation framework for aerial images. Starting from a neural network designed for semantic segmentation purpose, it consists in training this network to exploit user annotation inputs. At testing time, user annotations are input in the neural network without changing the parameters of the model, hence the interactive semantic segmentation process is swift and efficient.
Through experiments on two public aerial datasets, we have shown that interactive refinement is efficient for all classes. It improves classification results by 4\% on average for 120 clicks and mainly, produces segmentation maps which are visually more rewarding. We have shown that our interactive process is efficient whatever the network backbone is. We have also investigated different representations of the annotations and have concluded that clicks positioned inside instances and encoded using distance transform carry the most meaningful information. In the future, we will further investigate class-dependent annotation encoding.%
\sloppy

{
	\begin{spacing}{0.832}
		\bibliography{bibliography} 
	\end{spacing}
}

\end{document}